\RequirePackage{amsmath}
\documentclass[runningheads]{llncs}
\usepackage{amsmath}
\usepackage{graphicx}
\usepackage{mathtools}
\usepackage{multirow}
\usepackage{color}
\usepackage[hidelinks]{hyperref}
\usepackage[tight,footnotesize]{subfigure}
\usepackage{epstopdf}
\AppendGraphicsExtensions{.tif}
\graphicspath{{./Figures/}}

\begin{document}
\title{A Deep Learning Framework for Wind Turbine Repair Action Prediction Using Alarm Sequences and Long Short Term Memory Algorithms}
%
\titlerunning{A Deep Learning Framework for Wind Turbine
Repair Action Prediction}
%
\author{Connor Walker\inst{1,2} \and
Callum Rothon\inst{1,2} \and
Koorosh Aslansefat\inst{1,2} \and
Yiannis Papadopoulos\inst{1,2} \and
Nina Dethlefs\inst{1,2}}
\authorrunning{C. Walker, C. Rothon, K. Aslansefat, Y. Papadopoulos, N. Dethlefs}
%
\institute{University of Hull, Hull HU6 7RX, UK \and
AURA CDT, Hull, UK
\email{auracdt@hull.ac.uk}\\
\url{https://auracdt.hull.ac.uk/}}
\maketitle              

\begin{abstract}
With an increasing emphasis on driving down the costs of Operations and Maintenance (O$\&$M) in the Offshore Wind (OSW) sector, comes the requirement to explore new methodology and applications of Deep Learning (DL) to the domain. Condition-based monitoring (CBM) has been at the forefront of recent research developing alarm-based systems and data-driven decision making. This paper provides a brief insight into the research being conducted in this area, with a specific focus on alarm sequence modelling and the associated challenges faced in its implementation. The paper proposes a novel idea to predict a set of relevant repair actions from an input sequence of alarm sequences, comparing Long Short-term Memory (LSTM) and Bidirectional LSTM (biLSTM) models. Achieving training accuracy results of up to 80.23$\%$, and test accuracy results of up to 76.01$\%$ with biLSTM gives a strong indication to the potential benefits of the proposed approach that can be furthered in future research. The paper introduces a framework that integrates the proposed approach into O$\&$M procedures and discusses the potential benefits which include the reduction of a confusing plethora of alarms, as well as unnecessary vessel transfers to the turbines for fault diagnosis and correction. 

\keywords{Condition-based Monitoring (CBM)\and Deep Learning (DL)\and Long Short-term Memory (LSTM)\and Offshore Wind Farm (OSW)\and Repair Action Prediction\and Supervisory Control and Data Acquisition (SCADA)}
\end{abstract}

\section{Introduction}
O$\&$M is currently the second largest sub-sector market within OSW \cite{Catapult_OM_2021} and is projected to rise to the largest sub-sector by 2050 \cite{Catapult_OM_2021}. This development leads to an increased interest into research both to drive down costs associated with O$\&$M, as well as the safety of alarm-based systems.\\
Currently, O$\&$M consists of three major methods: preventative, failure-based and condition-based monitoring\cite{Maintenance_strategy_Zhou_2019}. The latter of these is the method most relevant to the success of alarm systems and alarm sequencing prediction. CBM relies on the alarm systems currently in place within turbines as well as SCADA systems\cite{Condition_Monitoring_Maldonado-Correa_2020}. SCADA has become a standard installation in larger turbines offshore in recent years, which means that the data collected is increasing rapidly, thus creating potential for new performance benchmarks within machine learning (ML) models applied in this area \cite{SCADA_Verhelst_2022}. 

The alarms linked to typical SCADA systems for OSW turbines allow monitoring of almost all sub-components \cite{SCADA_Health_Du_2017}. This is not an easy task for a number of reasons. Firstly, there is a certain risk of alarm flooding, especially since alarms may cascade during disturbances, as one symptom of the disturbance follows another. Alarm flooding refers to the relationship between alarm sequences and is defined as “10 or more enunciated alarms within a 10-minute period per operator” \cite{Alarm_Floods_Beebe_2012}. Where one alarm sounds, it is likely to then trigger other alarms due to the close relationships between components’ behaviour and overall performance. Multiple activated alarms can often distract from the original fault, leading to more downtime on the site whilst diagnostic reports are produced\cite{SCADA_analysis_Wei_2022}. Secondly, systems can generate false alarms, i.e. alarms caused by sensor failures and not as a result of process disturbances. To cascading and false alarms, one can add alarms created during maintenance. False and maintenance alarms are not only confusing for operators that oversee the health and safety of the turbine, but can also confuse automated ML algorithms, that are trained on this data,  e.g. for the purposes of fault isolation or generation of repair actions \cite{AlarmFault_Alexios_2019}. 

Current standards across industries, including EEMUA-191 ~\cite{EEMUA_191} and ANSI/ISA-18.2 ~\cite{ISA_18_2}, detail the design, management and procurement of alarm systems as well as the alarm management specific to process industries\cite{cai2019process}. 
These standards are used as a foundation for improving alarm processing and prediction of likely repair schedules, but they don’t prescribe or enforce specific techniques that address the significant problems mentioned above.
To address these issues and to achieve appropriate fault isolation and ultimately repair action prediction, in this paper, we propose a novel approach utilising alarm sequences to predict repair actions accurately and efficiently. Our contributions are:

\begin{itemize}
    \item A DL based approach to predict repair actions from a sequence of alarms. The paper experiments with both LSTM and BiLSTM algorithms for comparison of performance on this problem.
    \item A conceptual framework to integrate the idea of repair action prediction into OSW farm O$\&$M procedures. 
    \item The proposed use of reinforcement learning in a human-in-the-loop procedure to improve the accuracy of the DL model over time.
\end{itemize}

In section 2, we discuss the research question. In section 3, we detail our methodology and compare it to other approaches within the domain. The methodology section discusses the pre-processing of data, the design of the neural network, and experiments. Section 4 discusses results and application to industry and conclusions follow in Section 5.

\section{Research Questions}
\begin{figure*}[hbt!]
\centering
\includegraphics[width=\textwidth]{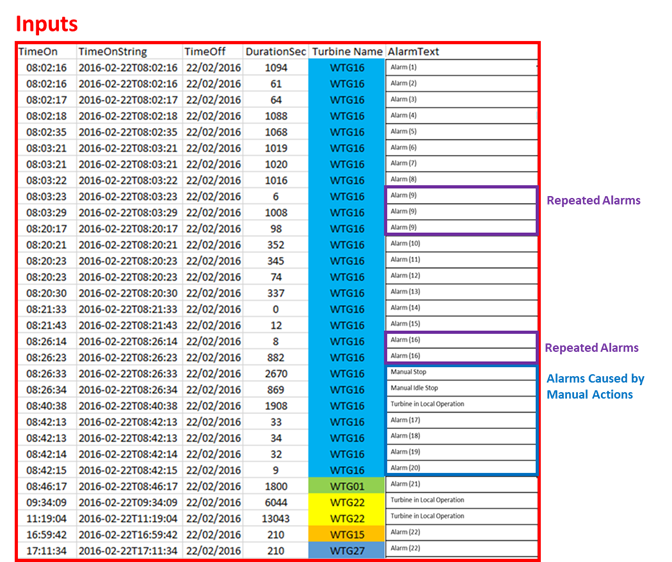}
\caption{A sample anonymised alarm Sequences for a specific date}
\label{fig_alarms}
\end{figure*}

Fig \ref{fig_alarms} shows a sample anonymised alarm sequence for a specific date. The OSW farm operators have to check these alarms and then based on experience decide or predict what should be repaired to fix the alarm. Some companies like SIEMENS provide a private and expensive software that can predict a potential required repair action(s) based on existing alarms for specific types of turbines. Based on our observation on Teesside Wind Farm, there are some specific dates that the operator faced more than 500 simultaneous alarms in a single day. This situation is challenging for human operators and, ideally, requires a technology that can assist operators by converting the large number of alarms to a small number of suggested required repair actions.

\begin{figure*}[hbt!]
\centering
\includegraphics[scale=0.7]{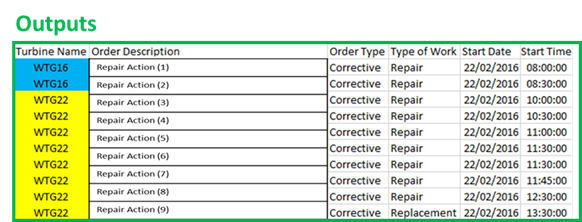}
\caption{A sample anonymised repair actions for a specific date}
\label{fig_repairs}
\end{figure*}

Fig \ref{fig_repairs} shows a sample of anonymised repair actions performed the same date in response to the alarms of Fig \ref{fig_alarms}. If we consider the alarm sequences observed as an input and repair actions as an output, our main research question is how a DL model can be used to predict the required repair actions given the input SCADA alarm sequences. Moreover, we ask how this model can be integrated into the operation and management processes of industry? and how it can evolve and improve itself over time.  

\section{Methodology} 
We address those research questions by developing DL models that perform this function based on Recurrent Neural Networks (RNN) and we show how these models can evolve and improve themselves over time using reinforcement learning.  It has been shown that RNN models have great promise in
Natural Language Processing (NLP) \cite{de2015survey}. An RNN is a neural network with feedback connections. Each cell's output not only gives information to the next layer, but it also gets feedback from itself. The Long short-term memory (LSTM) network is a kind of RNN that has one input layer, one output layer, and two hidden layers. We have chosen to base our methodology on  LSTM and BiLSTM models for the following advantages that they offer:  

\begin{itemize}
\item A DL based approach to predict repair actions from a sequence of alarms. The paper experiments with both LSTM and BiLSTM algorithms for comparison of performance on this problem
\item A conceptual framework to integrate the idea of repair action prediction into OSW farm O$\&$M procedures.  
\item The proposed use of reinforcement learning in a human-in-the-loop procedure to improve the accuracy of the DL model over time.
\end{itemize}

The LSTM model is commonly applied in NLP, to predict the most probable response to a given sequence of inputs. In this work, inputs are the alarm sequences and responses are repair actions. LSTMs are a form of Recurrent Neural Network, that have been proven effective in time-sequence forecasting, making them well suited to this application. Word Embeddings were used to prepare the alarm and response sequences for input into the LSTM layer. Word Embeddings are a fixed-length vector representation of words \cite{https://doi.org/10.48550/arxiv.1901.09069}, which provide context by situating the sequence inside a multi-dimensional semantic vector space. The embedded sequences are suitable for use as a numerical input into an LSTM layer. In this case, each alarm and response are embedded as vector.

Research by Cai et al. \cite{cai2019process} uses an LSTM-based process to predict the next alarm in a sequence given the previous alarm, with the intention of reducing alarm nuisance and overload. We use a similar LSTM-based methodology, with the crucial difference that we predict the most probable repair response to a sequence of alarms. This has multiple benefits over the mere prediction of the next alarm in a sequence, as the operator is given advice on repair. The operator may accept or reject the suggested repair and does not have to process the alarm information in order to infer the appropriate response. This automated support allows faster response to an alarm, which allows more rapid and dynamic O$\&$M, with reduced downtime of turbines while awaiting maintenance. Also, the likelihood of human error in the prescription of a response is reduced, provided that the system is trained to provide sufficient accuracy.

\begin{figure}[ht]
\includegraphics[width=\textwidth]{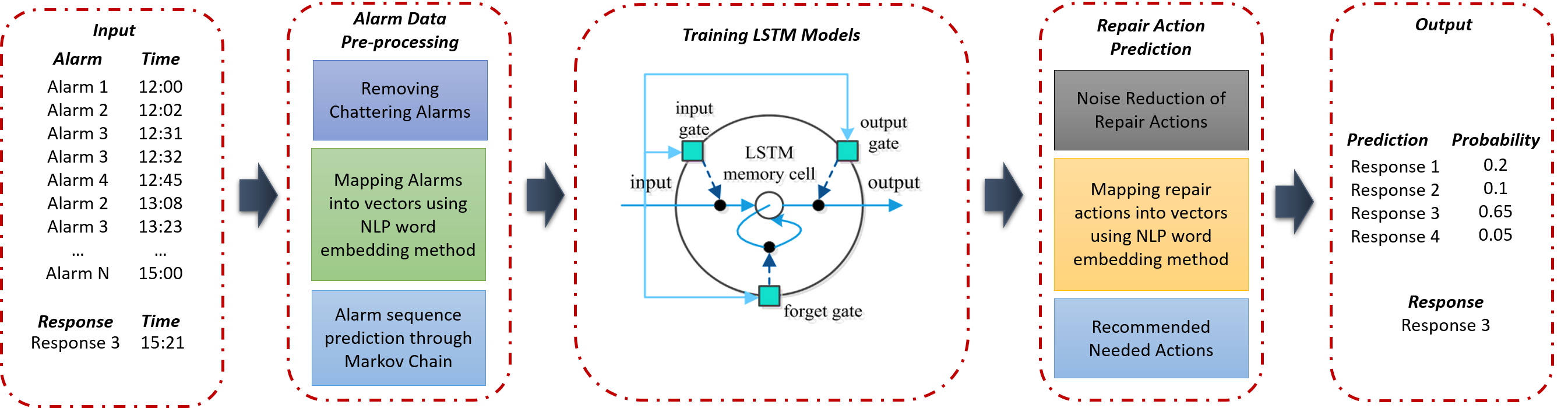}
\centering
\caption{ Flowchart of the methodology, including examples of inputs and outputs }
\label{fig6}
\end{figure}

The complete methodology is shown with examples of inputs and outputs in Fig \ref{fig6}. The inputs are received as a list of alarms in a time sequence, leading up to a response. These alarms then go through the pre-processing stages. Chattering alarms such as the two instances of Alarm 3 within a minute are removed, the alarms are embedded as vectors, and the alarm sequence is predicted through a Markov chain. The sequences of vectors are then input into the LSTM, with the response as the target during training. The response actions are processed to remove nuisance, mapped into vectors, and then output as recommendations. The outputs are  in the form of predicted response actions with a probability, and the most likely response is taken as the prediction. Each stage of the methodology is described in more detail throughout Section 3.

\subsection{Data Preparation and Pre-processing}
\begin{figure*}[hbt!]
\centering
\includegraphics[scale=0.3]{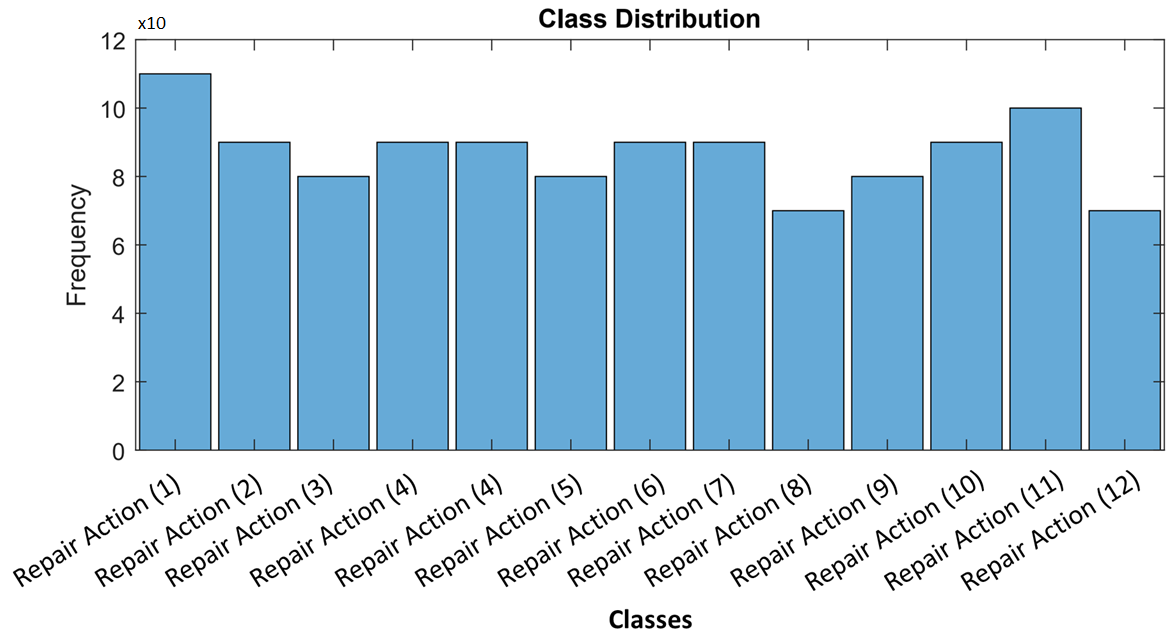}
\caption{Class distribution of anonymised repair actions (filtering repair actions with distribution less than 70)}
\label{fig3}
\end{figure*}

The Teesside OSW Farm was used as a case study to develop the method. This wind farm is known as Redcar Wind Farm and it is owned by EDF Energy. This OSW farm has 27, 2$-$3 MW turbines that can guarantee the total capacity of 62 MW.
In collaboration with EDF Energy R$\&$D London, we have received 5$-$years (from 2013 to 2018) of data in various forms such as alarm data, repair actions, repair plans, maintenance procedure, etc. Due to the confidentiality of the data, the results that can be shown in this section are limited and the provided data is anonymised. Fig \ref{fig3} shows the class distribution of anonymised repair actions. In this figure, repair actions that had a frequency of less than 70 were filtered. 

The pre-processing consisted of importing data into MATLAB, and preparing them for input into the LSTM network. The input data was a set of alarms and a set of responses, with each alarm and response having a time and text. This data was normalised and used to build sequences of alarms leading up to a response in a time series.

The data types were set, with the alarm and response times set as "datetime" and alarm and response texts were set as strings. Data cleaning was performed to remove unnecessary data and to remove punctuation from the alarm and response text. The variables were then imported into MATLAB.

The data was pre-processed for each turbine, with an identical process used for turbines 1 through to 27, with turbine 10 being held back for testing. The “TimeOnString” was initialised for the alarm and response files, and a variable “mem” defined as 20, which defines the time interval over which an alarm and response can be associated. In this case, a response can be associated with an alarm that occurred up to 20 days previously, which was considered a reasonable timescale. 

The training data was accumulated from the alarm and response schedules from turbines 1-27 over a period of 20 days, by iteratively passing over them and collecting alarm-response pairs. 
 This was completed first for alarms 1-27. From this 27 timetables TT1-27 will have been generated, containing the responses and their preceding alarms. It must be noted at this time that causality has not yet been established (or more correctly, predicted), and the association is purely through the alarm occurring prior to the action.

Some further data cleaning was then performed on the timetables. Noise characters were specified and removed, and empty strings are deleted. Responses with less than 2 entries are classed as infrequent and were removed from the training dataset, as they are unlikely to contribute meaningfully to the model. This threshold was set by plotting a histogram and inspecting the data empirically. 

\begin{figure*}[hbt!]
\centering
\includegraphics[scale=0.12]{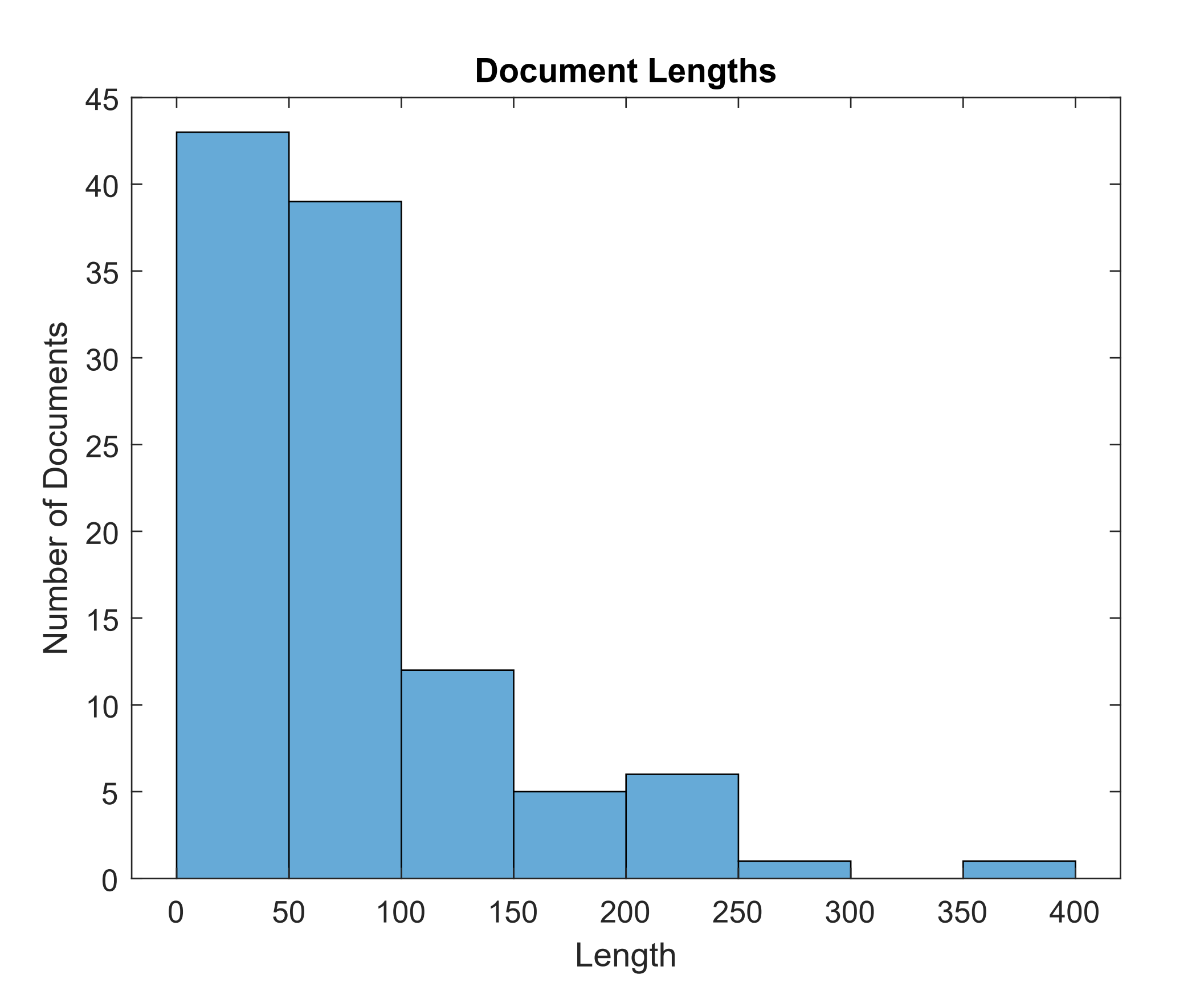}
\caption{Histogram of document length}
\label{fig4}
\end{figure*}

The data was split into a training and validation set, using a holdout of 0.3, which was determined to be a suitable validation split. From the validation set, a further holdout of 0.5 was used to form the testing data , resulting in a training data set of 70$\%$ of the data, and validation and testing data sets of 15$\%$ each. The text data and the labels are then extracted from these partitioned data sets and labelled as “TextData” and “Y”. A word cloud of the text training data was plotted for some initial explainability, demonstrating the distribution of keywords in the data set, showing that certain keywords were prevalent. The keywords cannot be discussed here due to the confidentiality of the dataset. The text data sets were then converted to lowercase characters, tokenised, and punctuation was erased. The benefits of pre-processing via tokenisation are outlined in \cite{camacho2017role}, which demonstrates that simple tokenisation outperforms more complex pre-processing.  The data was then encoded using seq2seq \cite{sutskever2014sequence}, converting the documents into sequences.
Fig \ref{fig4} illustrates the histogram of the document length. Based on the distribution of the document lengths, a target length of 75 was used.

\subsection{Building the LSTM Network}
Unlike \cite{cai2019process}, the LSTM network presented in this paper aimed to determine the most likely response (repair actions) to a sequence of alarms as opposed to the next alarm in the sequence. In this methodology, both conventional LSTM and Bi-Directional LSTM (BiLSTM) layers were trialled. When using an LSTM layer, the sequence runs from start to finish, while the BiLSTM runs start to finish and finish to start, and is generally considered to give improved accuracy of predictions, as shown in \cite{cui2018deep}, and has been utilised in NLP by \cite{BiLSTMKey}. The structure of the LSTM model is shown in Fig \ref{fig2}, both during training and during prediction.

To embed the words into vectors for input into the LSTM layer, Word2vec was used with pre-trained embeddings. Pre-trained embeddings, a form of transfer learning, are trained using wide corpuses, and give more accurate embeddings than when training from scratch. \cite{10.1007/s10664-022-10118-5} recommends careful consideration of the pre-trained embeddings that are used, as the context of the training will impact how applicable the embeddings are for use in other models.

The layers of the model were defined with MATLAB variables used to set the parameters of the model which allowed for adjustment where required. The structure of the model is shown during training in Fig \ref{fig2}a and during prediction in Fig \ref{fig2}b.

The sequence input layer receives the sequences created in the pre-processing, and is defined as being dependent upon the input size variable, resulting in a one-dimensional input layer, corresponding to the vectorised inputs. The word embedding layer is dependent upon the embedding dimension and the number of hidden units as defined in the variables, giving an embedding layer with 300 dimensions and as many hidden units as there are unique words in the data set. The embedding layer converts the words input into vectors, allowing the LSTM/BiLSTM to process them as numerical data. 

The LSTM/BiLSTM layer was defined by the hidden size, giving 300 dimensions, and the output mode defined as “last” which will output the last step of the sequence, the predicted response. This differs from more conventional use of an LSTM, which will aim to predict the next step in the sequence. In this case, the output will be a prediction of the most probable response action given an input sequence. During training, the LSTM/BiLSTM layer will receive the response action at the end of the sequence as the target, and will train based on this. 

The fully connected layer is dependent upon the number of elements present in “YTrain”. A softmax layer bounds the outputs between 0 and 1, and a classification layer completes the model, delivering predictions between 0 and 1 for the responses. The response with the highest probability is taken as the predicted response.

The LSTM network receives the vectorised sequences as the inputs, with the responses as the targets, and outputs a state which is passed onto the next step of the LSTM.  The LSTM trains over a set number of epochs, with an Adam optimiser used. Adam was demonstrated to be more effective than Root Mean Squared Propagation (RMSProp) in many applications by \cite{shaziya2020optimization}, hence its use here. The model is validated using the 15$\%$ of data partitioned as the validation set. After the full number of epochs, a trained model is yielded, and the model can be tested and predictions generated from new data. 

\subsection{Training of the LSTM Network}
When training the LSTM, the training parameters were considered to optimise the performance of the model. The number of epochs was set at 50, which was determined to give acceptable accuracy. The gradient threshold was set at 1, and the initial learning rate was set at 0.01. A training progress plot was enabled to give an insight into how the training was progressing. The architecture of the network during training is shown in shown Fig \ref{fig2}a, with the alarm sequence documents as inputs that undergo embedding, and the vectorised response documents as the targets for training.

\subsection{Testing and Prediction} 
The testing involved using the partitioned 15$\%$ of the data set reserved for testing, which has previously not been processed by the model. Prior to testing, the data was pre-processed in the same manner as the training and validation data,  predictions were generated from the reserved data, and the accuracy was calculated based on the number of correct predictions divided by the total number of predictions.

\begin{figure}[htbp]
\includegraphics[scale=0.373]{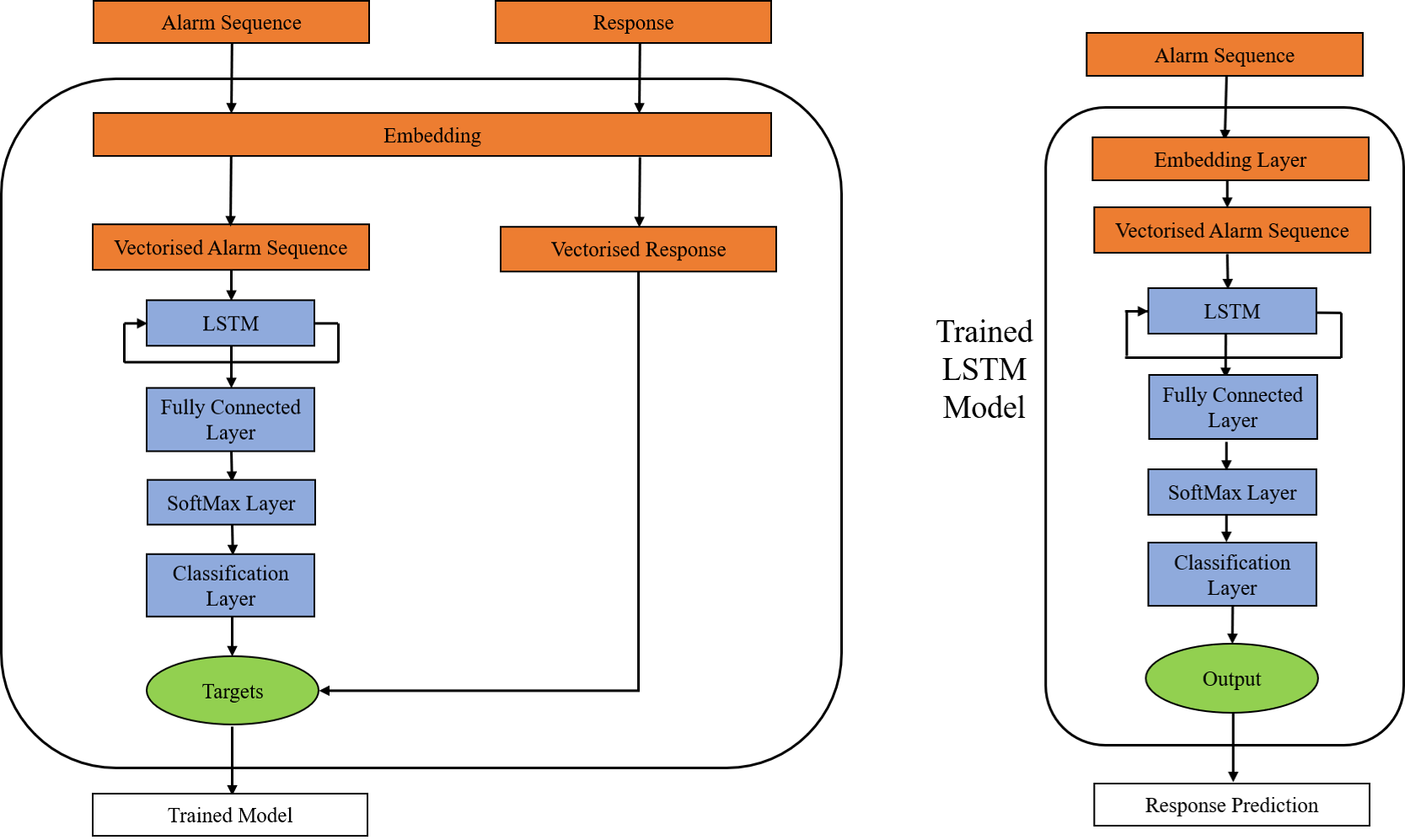}
\centering
\caption{a. Architecture of the model during training, b.Architecture of the trained model during testing. }
\label{fig2}
\end{figure}

Testing of the LSTM network was also performed with the reserved data from Turbine 10. This testing involved the input of the previously unseen data into the trained model to generate predictions. The data was pre-processed as with the data for the previous turbines, and the sequenced alarms were fed into the model, and then predicted responses and scores were produced. Based on these labels, the new reports were produced. The architecture of the trained model during testing and prediction is shown in Fig \ref{fig2}b, with the alarms providing inputs into the trained model, and a response prediction being the output.

\section{Results and Discussion}

\subsection{Results}
    Table \ref{tab1} provides the prediction accuracy of both LSTM and BiLSTM for train and test data sets.
    \begin{table}[htbp]
      \centering
      \caption{Prediction accuracies of LSTM vs. BiLSTM}
        \begin{tabular}{lrr}
        \hline
                 & \multicolumn{1}{l}{Training Accuracy} & \multicolumn{1}{l}{Test Accuracy} \\
        \hline
        LSTM     & 78.65 & 64.70 \\
        BiLSTM   & 80.20 & 75.88 \\
        \hline
        \end{tabular}%
      \label{tab1}%
    \end{table}%

\begin{figure}[htbp]
\includegraphics[scale=0.6]{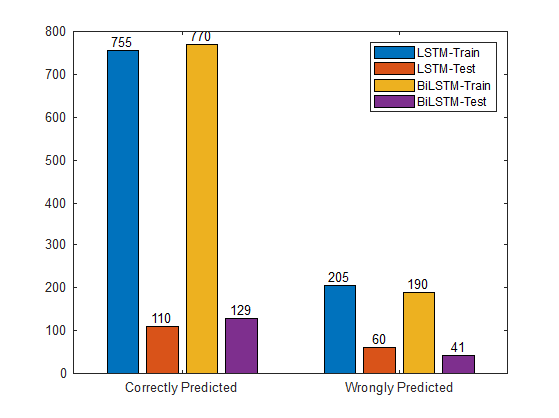}
\centering
\caption{ Plot of the correctly and incorrectly predicted results for the LSTM and BiLSTM models during testing and training. }
\label{fig7}
\end{figure}

    The LSTM and BiLSTM models yielded different prediction accuracies, with the BiLSTM delivering superior accuracy of 75.88$\%$, compared to the 64.70$\%$ delivered by the LSTM. This accuracy is slightly lower than the results of 78-81.4$\%$ stated by \cite{cai2019process}, but the accuracies are comparable, and it is likely that with an improved data set and some further optimisation of the model that accuracy in the region of 90$\%$ could be achieved in future. It should be noted that the model presented by \cite{cai2019process} predicts the next alarm in a sequence, not repair actions, so the differing aims of the papers must be considered when drawing comparisons. 
    To give further insight into the accuracy of the models, a bar chart  in Fig \ref{fig7} plots the number of correct and incorrect predictions made by the LSTM and BiLSTM models during testing and training.

It is shown that the BiLSTM outperformed the one-directional LSTM narrowly during testing with more  correct and less false predictions. Importantly, the BiLSTM outperformed the LSTM considerably during testing, decreasing false predictions by almost a third. Any occurrences of unnecessary or missed responses could cause issues for industry, with an unnecessary repair operation causing inconvenience and unnecessary expenditure, and a missed repair action allowing a fault to develop. Therefore, the use of a BiLSTM model is recommended in this application. 

\subsection{Benefits for Industry}
The successful implementation of a response prediction network based on alarm sequences would have some implications for O$\&$M planning in the OSW industry. As the proposed system considers alarms that precede a response action by up to 20 days, it is possible that a response can be predicted with fewer alarms than are currently required. While it is unlikely that an accurate response could be predicted from a single alarm, reasonably accurate predictions may be made in advance of a failure.

The prediction of a response action prior to an alarm would provide major benefits, as it would be possible to plan and undertake predictive maintenance based on early warning from alarms before a major alarm or failure occurs. The benefits of predictive maintenance would include a reduction of turbine downtime while a response is planned and implemented, and a reduction in the need for human decision-making (and therefore human error) in planning complex response. While it is unlikely that the need for a human in the process would be completely removed, supported decision-making would provide benefits for O$\&$M planning.

It is also likely that if a response can be predicted from a shorter sequence of alarms, then it may allow issues to be rectified before they progress to a more serious, and therefore more costly issue. 
\subsection{Integration into Industry}
\begin{figure}[htbp]
\includegraphics[scale=0.3]{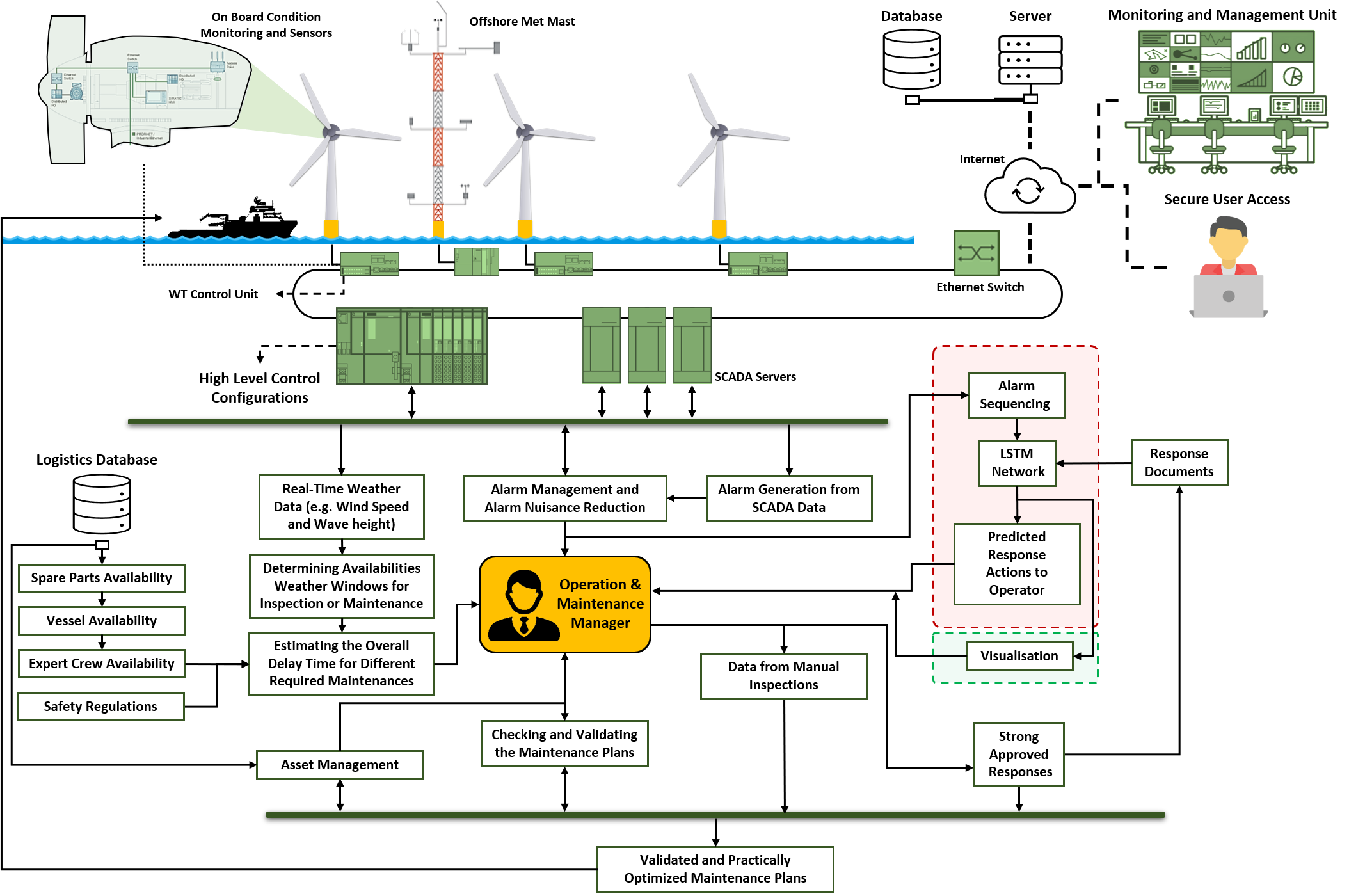}
\centering
\caption{ System diagram of the proposed integration of the repair action recommendation network into the existing O$\&$M alarm system. }
\label{fig8}
\end{figure}

In order for the OSW Industry to benefit from the use of this model, its implementation into the existing O$\&$M alarm systems must be considered. A diagram  showing the recommended integration of the repair action recommendation network is shown in Fig \ref{fig8}. In the diagram, the repair action recommendation network is bounded by an orange box, with the LSTM network corresponding to the trained network shown in Fig \ref{fig2}b. The addition of visualisation for explainability which is considered as a potential future development, is bounded in a green box.

The system receives alarms generated from SCADA data, which will be sequenced, and input into the LSTM network, then output response predictions to the O$\&$M manager for approval. We are currently developing an extension of this approach to add explainability in the recommendations of the LSTM, as this will allow improved understanding of the recommendations by the O$\&$M manager.

\begin{figure}[htbp]
\includegraphics[scale=0.4]{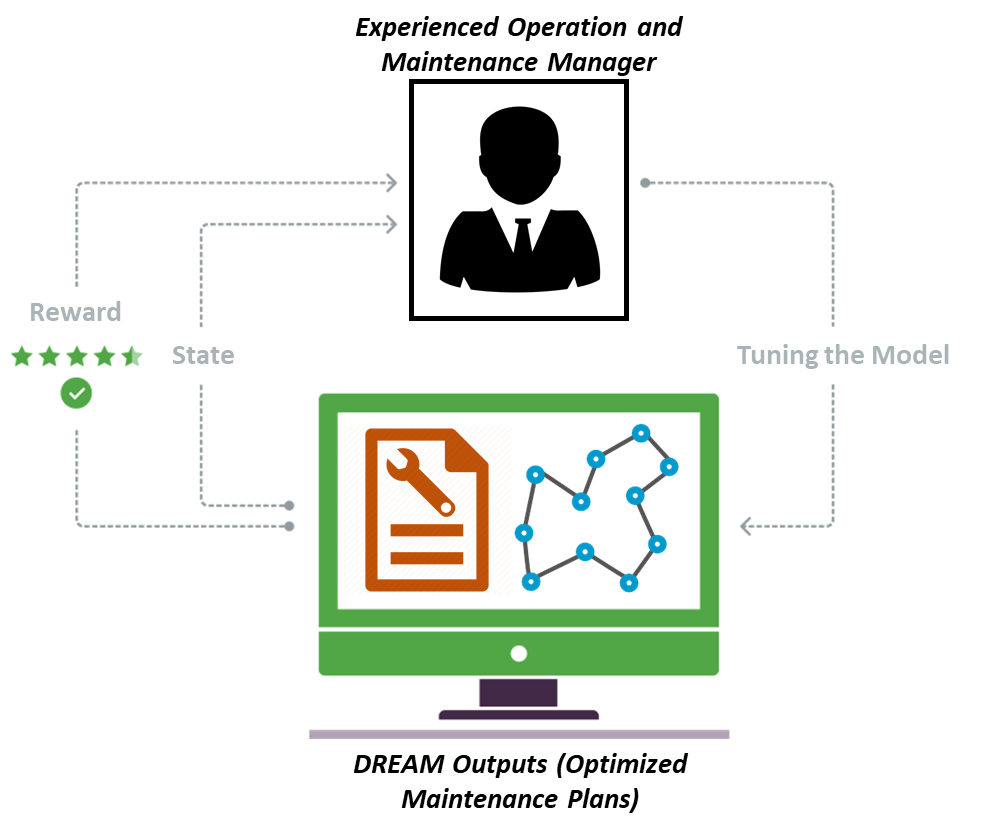}
\centering
\caption{A proposed procedure for using reinforcement learning-like technique and use human-in-the-loop scoring of recommendations to improve the system's performance over time.}
\label{fig9}
\end{figure}

It is unlikely that human involvement can be completely removed from the system. Therefore, it is recommended that the repair action recommendation framework is subordinate to the O$\&$M Manager who has ultimate authority. Daily, the repair recommendation network can provide suggestions for the repair actions based on the existing alarms, and then the O$\&$M manager reviews the outputs, choosing to accept or reject the recommendations and rating them. In a case of having low ratings, the system will change some parameters, update itself, and generate another output. The algorithm will run until the results reach a certain level of acceptance. As an example, if the O$\&$M manager put a low rate for the recommendation, then the application asks the correct needed action and updates itself accordingly. Fig \ref{fig9} illustrates the overall idea of using reinforcement learning to make the repair action recommendation more realistic. 
\subsection{Current Limitations}
A potential limitation in the system is the varying terminology used in alarms generated from SCADA data. This may lead to misidentification of alarms, which would result in reduced accuracy of predictions. This in turn could yield incorrect predictions, or possibly false negatives, i.e. situations where the system fails to predict a response when one is required. In order to remedy this limitation, it would be necessary to either expand the training data set to include all terminologies used, or to train the network for each set of terminology. As a larger data set would lead to increased training time and computational cost, it would be more realistic to train the LSTM for each set of alarms, which would differ between operators.

Another potential limitation of the system is that it depends on alarms and responses having enough occurrences to meaningfully contribute to the training. During the training of the model, any alarms with fewer than 2 occurrences were considered as too infrequent to meaningfully contribute to the training, and so were removed from the data set. This potentially leads to certain alarms being ignored, which could prevent more accurate predictions from being made, and the alarms could relate to major issues. One solution to this would be to use a larger data set, gained over a longer time period. However, this would cause increased computational cost during training, and long-term alarm data from the industry is largely unavailable.

\section{Conclusions and Future Work}
This paper proposed a novel approach to repair action prediction using RNNs. Both LSTM and BiLSTM models have been trained and tested to predict repair actions with the input of alarm sequences. Using 5 years of alarm sequence data  by EDF Energy, we have successfully shown the potential of these networks to produce beneficial contributions to the domain by achieving test accuracy’s of up to 76$\%$. The results achieved using this DL approach have proved comparable with other research applied to relevant but different problems. With a lack of usable clean data being a major factor in achieving high prediction accuracy, it has been noted that the DL approach may have limitations in realistic applications  with the current quality of data. However, one solution that has been proposed as future work is to use reinforcement learning with a human-in-the-loop procedure. 

In the future, we shall use information in the data provided by EDF to distinguish between minor and major repairs. Explainability approaches will  be integrated to the trained model to make its behaviour interpretable for the operators. Methods like SafeML \cite{aslansefat2020safeml, aslansefat2021toward} will  be added on top of the trained model to improve the reliability and safety of the prediction at runtime. Finally, we will experiment using DL to auto-create  a temporal casual model like a dynamic fault tree that can illustrate the causality between alarm sequences. Such a model could help the system’s designers and operators to gain more insights regarding the alarm sequence behaviour \cite{aslansefat2020performance,simeu2011methodology}.

\section*{Code Availability}
\label{CA}
Regarding the research reproducibility, codes and functions supporting this paper are published online at GitHub: 
\\
\href{https://github.com/koo-ec/OWF_Repair_Action_Recommender}{https://github.com/koo-ec/OWF$\_$Repair$\_$Action$\_$Recommender}.\\
Due to the confidentiality of data provided by EDF Energy, it is not possible for the data set to be made public.

\section*{Acknowledgement}
This work was supported by the Secure and Safe Multi-Robot Systems (SESAME) H2020 Project under Grant Agreement 101017258. We would like to thank EDF Energy R$\&$D UK Centre, AURA Innovation Centre and University of Hull for their support.

\bibliographystyle{splncs04}

\begin{thebibliography}{10}
\providecommand{\url}[1]{\texttt{#1}}
\providecommand{\urlprefix}{URL }
\providecommand{\doi}[1]{https://doi.org/#1}

\bibitem{https://doi.org/10.48550/arxiv.1901.09069}
Almeida, F., Xexéo, G.: Word embeddings: A survey (2019).
  \doi{10.48550/ARXIV.1901.09069}, \url{https://arxiv.org/abs/1901.09069}

\bibitem{aslansefat2020performance}
Aslansefat, K., Gogani, M.B., Kabir, S., Shoorehdeli, M.A., Yari, M.:
  Performance evaluation and design for variable threshold alarm systems
  through semi\-markov process. ISA transactions  \textbf{97},  282--295 (2020)

\bibitem{ISA_18_2}
of~Automation, I.S.: {Management of Alarm Systems for the Process Industries}.
  Standard, International Society of Automation, North Carolina, United States
  (2016)

\bibitem{BiLSTMKey}
Basaldella, M., Antolli, E., Serra, G., Tasso, C.: Bidirectional lstm recurrent
  neural network for keyphrase extraction. In: Italian research conference on
  digital libraries. pp. 180--187. Springer (2018)

\bibitem{cai2019process}
Cai, S., Palazoglu, A., Zhang, L., Hu, J.: Process alarm prediction using deep
  learning and word embedding methods. ISA transactions  \textbf{85},  274--283
  (2019)

\bibitem{camacho2017role}
Camacho-Collados, J., Pilehvar, M.T.: On the role of text preprocessing in
  neural network architectures: An evaluation study on text categorization and
  sentiment analysis. arXiv preprint arXiv:1707.01780  (2017)

\bibitem{cui2018deep}
Cui, Z., Ke, R., Pu, Z., Wang, Y.: Deep bidirectional and unidirectional lstm
  recurrent neural network for network-wide traffic speed prediction. arXiv
  preprint arXiv:1801.02143  (2018)

\bibitem{de2015survey}
De~Mulder, W., Bethard, S., Moens, M.F.: A survey on the application of
  recurrent neural networks to statistical language modeling. Computer Speech
  and Language  \textbf{30}(1),  61--98 (2015)

\bibitem{10.1007/s10664-022-10118-5}
Ding, Z., Li, H., Shang, W., Chen, T.H.P.: Can pre-trained code embeddings
  improve model performance? revisiting the use of code embeddings in software
  engineering tasks. Empirical Software Engineering  \textbf{27}(3) (2022).
  \doi{10.1007/s10664-022-10118-5},
  \url{https://doi.org/10.1007/s10664-022-10118-5}

\bibitem{SCADA_Health_Du_2017}
Du, M., Yi, J., Mazidi, P., Cheng, L., Guo, J.: A parameter selection method
  for wind turbine health management through scada data. Energies
  \textbf{10}(2), ~253 (2017)

\bibitem{Alarm_Floods_Beebe_2012}
Dustin~Beebe, Steve~Ferrer, D.L.: Alarm floods and plant incidents.
  \url{https://www.digitalrefining.com/article/1000558/alarm-floods-and-plant-incidents#.YkLrZefMIuV}
  (2012), online; Accessed 27 March 2022

\bibitem{EEMUA_191}
Equipment, E., Association, M.U.: { EEMUA Publication 191 Alarm systems - a
  guide to design, management and procurement }. Standard, Engineering
  Equipment and Materials Users Association, London, UK (Mar 2019)

\bibitem{AlarmFault_Alexios_2019}
Koltsidopoulos~Papatzimos, A., Thies, P.R., Dawood, T.: Offshore wind turbine
  fault alarm prediction. Wind Energy  \textbf{22}(12),  1779--1788 (2019).
  \doi{https://doi.org/10.1002/we.2402},
  \url{https://onlinelibrary.wiley.com/doi/abs/10.1002/we.2402}

\bibitem{Condition_Monitoring_Maldonado-Correa_2020}
Maldonado-Correa, J., Martín-Martínez, S., Artigao, E., Gómez-Lázaro, E.:
  Using scada data for wind turbine condition monitoring: A systematic
  literature review. Energies  \textbf{13}(12) (2020).
  \doi{10.3390/en13123132}, \url{https://www.mdpi.com/1996-1073/13/12/3132}

\bibitem{Catapult_OM_2021}
{Offshore Renewable Energy (ORE) Catapult}: {Offshore Wind Operations and
  Maintenance, A £9 Billion per year opportunity by 2030 for the UK to Seize}.
  \url{https://ore.catapult.org.uk/wp-content/uploads/2021/05/Catapult-Offshore-Wind-OM_final-050521.pdf}
  (2021), online; Accessed 29 March 2022

\bibitem{simeu2011methodology}
Simeu-Abazi, Z., Lefebvre, A., Derain, J.P.: A methodology of alarm filtering
  using dynamic fault tree. Reliability Engineering and System Safety
  \textbf{96}(2),  257--266 (2011)

\bibitem{sutskever2014sequence}
Sutskever, I., Vinyals, O., Le, Q.V.: Sequence to sequence learning with neural
  networks. In: Advances in neural information processing systems. pp.
  3104--3112 (2014),
  \url{https://papers.nips.cc/paper/5346-sequence-to-sequence-learning-with-neural-networks.pdf}

\bibitem{SCADA_Verhelst_2022}
Verhelst, J., Coudron, I., Ompusunggu, A.P.: Scada-compatible and scaleable
  visualization tool for corrosion monitoring of offshore wind turbine
  structures. Applied Sciences  \textbf{12}(3) (2022).
  \doi{10.3390/app12031762}, \url{https://www.mdpi.com/2076-3417/12/3/1762}

\bibitem{SCADA_analysis_Wei_2022}
Wei, L., Qian, Z., Pei, Y., Wang, J.: Wind turbine fault diagnosis by the
  approach of scada alarms analysis. Applied Sciences  \textbf{12}(1) (2022).
  \doi{10.3390/app12010069}, \url{https://www.mdpi.com/2076-3417/12/1/69}

\bibitem{shaziya2020optimization}
Zaheer, R., Shaziya, H.: A study of the optimization algorithms in deep
  learning. In: 2019 third international conference on inventive systems and
  control (ICISC). pp. 536--539. IEEE (2019)

\bibitem{Maintenance_strategy_Zhou_2019}
Zhou, P., Yin, P.: An opportunistic condition-based maintenance strategy for
  offshore wind farm based on predictive analytics. Renewable and Sustainable
  Energy Reviews  \textbf{109}, ~1--9 (2019)

\bibitem{aslansefat2020safeml}Aslansefat, K., Sorokos, I., Whiting, D., Tavakoli Kolagari, R. \& Papadopoulos, Y. SafeML: safety monitoring of machine learning classifiers through statistical difference measures. {\em International Symposium On Model-Based Safety And Assessment}. pp. 197-211 (2020)

\bibitem{aslansefat2021toward}Aslansefat, K., Kabir, S., Abdullatif, A., Vasudevan, V. \& Papadopoulos, Y. Toward improving confidence in autonomous vehicle software: A study on traffic sign recognition systems. {\em Computer}. \textbf{54}, 66-76 (2021)

\end{thebibliography}

\end{document}